\documentclass[10pt]{article}
\usepackage[margin=0.9in]{geometry}
\usepackage{hyperref}       
\usepackage{url}            
\usepackage{booktabs}       
\usepackage{amsfonts}       
\usepackage{nicefrac}       
\usepackage{microtype}      
\usepackage{bm}
\usepackage{graphicx}
\usepackage{subfigure}
\usepackage{amsmath}
\usepackage{floatrow}
\usepackage{bbold}

\usepackage[font=footnotesize,skip=0pt]{caption}
\usepackage{bbm}

\begin{document}\normalsize

\title{Fast Real-time Counterfactual Explanations}
\author{
  \small Yunxia Zhao, Department of Civil and Environmental Engineering\\
  \small University of California, Irvine, USA \\
  \small email: \texttt{yunxiaz1@uci.edu}
}

\date{}
\maketitle

\begin{abstract}
Counterfactual explanations are considered, which is to answer {\it why the prediction is class A but not B.} Different from previous optimization based methods, an optimization-free Fast ReAl-time Counterfactual Explanation (FRACE) algorithm is proposed benefiting from the development of multi-domain image to image translation algorithms. Built from starGAN, a transformer is trained as a residual generator conditional on a classifier constrained under a proposal perturbation loss which maintains the content information of the query image, but just the class-specific semantic information is changed. The transformer can transfer the query image to any counterfactual class, and during inference, our explanation can be generated by it only within a forward time. It is fast and can satisfy the real-time practical application. Because of the adversarial training of GAN, our explanation is also more realistic compared to other counterparts. The experimental results demonstrate that our proposal is better than the existing state of the art in terms of quality and speed.
\end{abstract}

\section{Introduction}

\label{sec:intro}

Although deep learning systems have been widely applied in computer vision tasks~\cite{he2016deep,ren2015faster,wang2017idk,Wang_2018_ECCV,goodfellow2014generative}, the black-box nature of it is still unopened and remains mysterious.
This hindered its deployment in real-world applications, because generally speaking, it is hard to trust an AI system if it can not justify its decision. Motivated by this requirement, explainable AI(XAI), which aims to unravel the mystery of network prediction, has attracted more and more attention in recent years. The dominant technology in computer vision is attribution~\cite{shrikumar2017learning,sundararajan2017axiomatic,selvaraju2017grad}, which produces a saliency map for each query image given a pre-trained model. The high value on the map highlights the image regions 
that is mainly responsible for a certain prediction. This way of attribution meets the requirements of simple visualization. However, it lacks interaction in practical applications. For instance, when a lesion is predicted to be A, a doctor would naturally ask ``why is the diagnosis A rather than B?'' A similar question would be proposed by a child who is learning to recognize different characters. When the tutoring system tells her that the character displayed is ``E'', the child may ask ``why not ``F''?'' In this case, we need an  interpretation mechanism that can interact with user in real time.

Counterfactual explanations~\cite{goyal2019counterfactual,wang2020scout} (or contrastive explanations in some literature~\cite{dhurandhar2018explanations}) were introduced to solve this problem. Basically, the explanation is usually implemented as ``correct class is A (prediction class). If it is class B (counterfactual class), some regions have to be changed as follows.'' Existing possible transformations include image perturbation~\cite{dhurandhar2018explanations} and image replacement~\cite{goyal2019counterfactual}. However, image perturbations frequently leave the space of natural images. 
The generated images are not realistic. It is so hard to convince users with such synthesized images. The current standard image replacement algorithms not only have the same problem, but also are too time-consuming for interactive applications due to its exhaustive search mechanism which is far too complex. Speed is critical for applications such as machine teaching~\cite{zhu2018overview,goyal2019counterfactual}, where explanation algorithms should operate in real-time, and ideally in low-complexity platforms such as mobile devices. \cite{wang2020scout} recently proposed SCOUT algorithm to find semantic corresponding discriminant features for two classes, but its results are hard to interpret, which is another problem with the practical application.

To address such problems, in this work, a Fast ReAl-time Counterfactual Explanation (FRACE) algorithm is proposed benefiting from the success of Generative Adversarial Network (GAN)~\cite{goodfellow2014generative} and its variants~\cite{mirza2014conditional,zhu2017unpaired,choi2018stargan}. Based on starGAN~\cite{choi2018stargan}, we use residual generator to train the transformation. Different from the standard generator, the residual generator attempts to generate a perturbation that causes the query image to be classified as counterfactual. In order to guarantee that the generated images are conditional on the classifier used to explain and there is a minimum change occurring from the query, a perturbation loss and this classifier are added to the training of the generator. Because our FRACE is based on GAN, the produced images are more realistic compared to previous \cite{dhurandhar2018explanations,goyal2019counterfactual}. Because of optimization-free during inference time, to generate our explanation is much fast. In experiments, we compare FRACE to state of the art and prove the effectiveness and efficiency of our proposals. 






\section{Counterfactual explanation generation}

\begin{figure}[t]
\vskip 0.2in
\begin{center}
\centerline{\includegraphics[width=0.7\columnwidth]{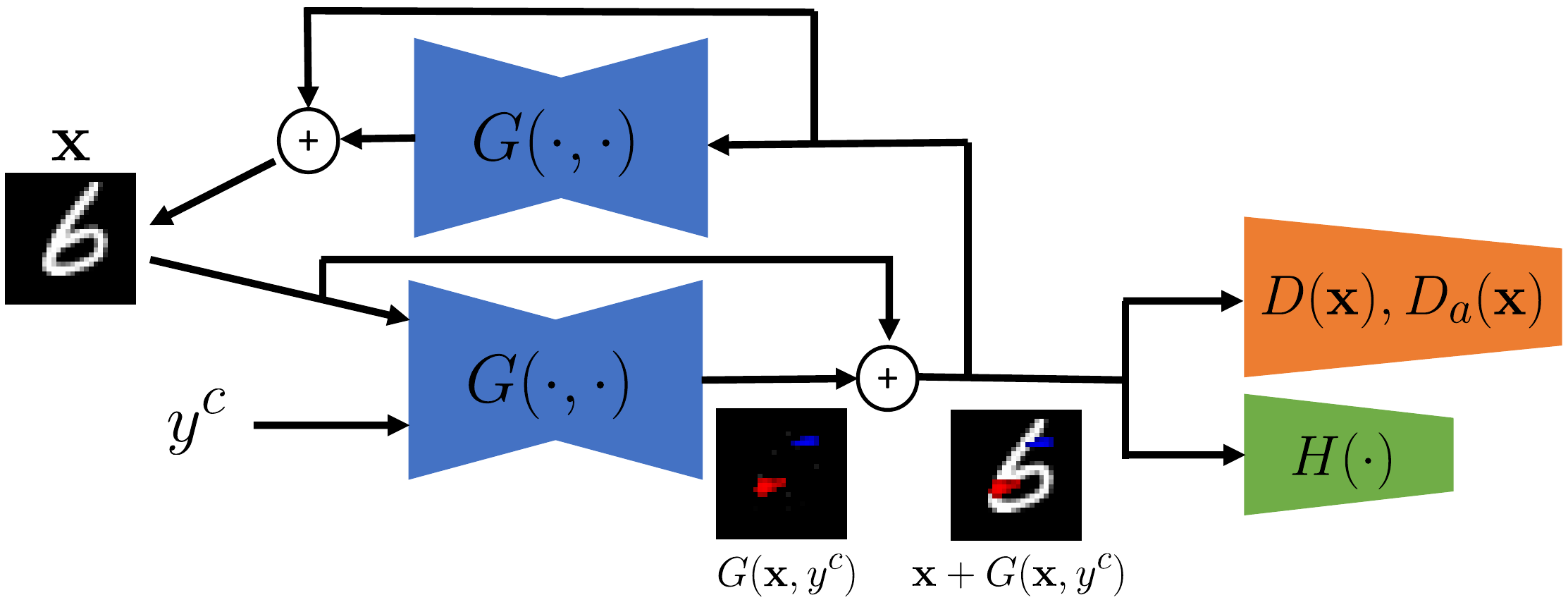}}
\caption{Architecture of our FRACE. In the example, the prediction class ($y^*$) and ground truth (y) of the query image is ``6'' and counterfactual class $y^c$ is ``5''. Our explanation would be ``If the red regions are erased and blue regions are added, the image would be ``5''.''}
\label{fig:arch}
\end{center}
\vskip -0.2in
\end{figure}
\begin{figure}[t]
\vskip 0.2in
\begin{center}
\centerline{\includegraphics[width=7.0cm]{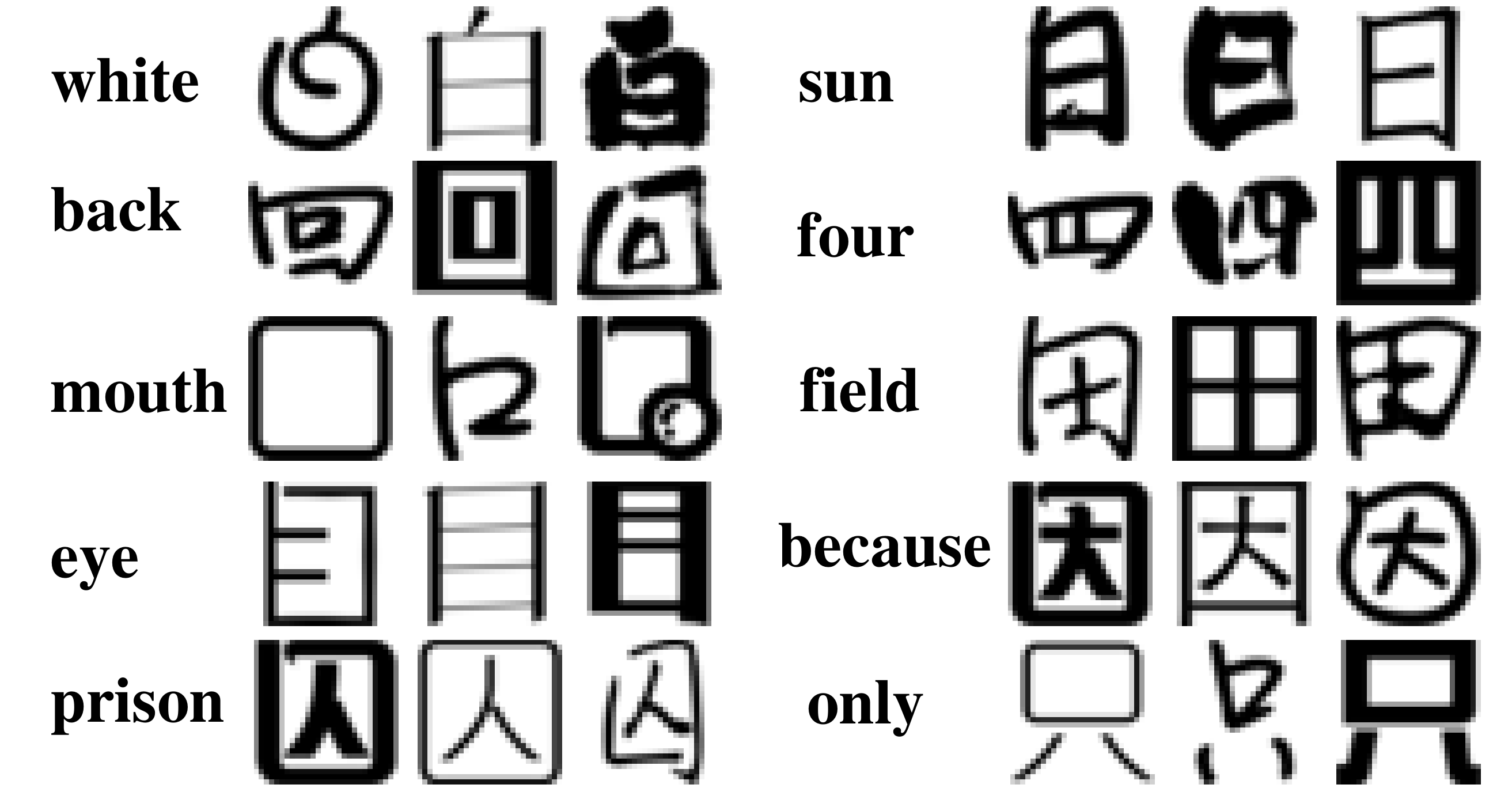}}
\caption{Chinese Character examples. The dataset consists of ten classes. Three examples are shown for each category.}
\label{fig:chinese_example}
\end{center}
\vskip -0.2in
\end{figure}

Our explanation works on a C-class classification problem, mapping image $\mathbf{x}$ with ground truth label $y$ to a scale label $y^*$, where $y, y^* \in \mathcal{Y} = \{1,2,...,C\}$, for a pre-trained classifier $H(\mathbf{x})$. $y^*$ is the classifier's prediction for $\mathbf{x}$. This classifier is the one to be explained. A user-choosing counterfactual class ($y^c$) is supposed to be given. For transformation based counterfactual explanation, the goal is to obtain a transformer $F$ such that $F(\mathbf{x}) \rightarrow \mathbf{x}^c$, where the $\mathbf{x}^c$ is the target explanation. In this paper, we train a generator by GAN to build the transformer, which can explain why the prediction of $H$ is $y^*$ but not $y^c$ by translating $\mathbf{x}$ to $\mathbf{x}^c$ under the condition of $y^c$.

We leverage the starGAN approach~\cite{choi2018stargan} to build our FRACE with some modifications. 
Basically, the goal is to train a single generator $G$ that learns mappings among multiple domains, where the domain is the class label. To achieve this, we train $G$ to translate an input image $\mathbf{x}$ with label $y$ to a perturbation towards $y^c$, such that $\mathbf{x} + G(\mathbf{x}, y^c) \rightarrow \mathbf{x}^c$. An auxiliary classifier~\cite{odena2017conditional}, $D_a$, is introduced to control the target domains. The classifier $H$ is followed by the generator to guarantee the produced images fit its distribution.\\
\noindent\textbf{Adversarial loss} The basic GAN adversarial loss is used to guarantee the generated images indistinguishable with real-world images by an associated discriminator via min-max optimization fashion. However, different from standard GAN, instead of learning the $G(\mathbf{x}, y^c)$ directly, we advocate to learn it by residual connection way considering 1) it makes the optimization easier, which already proven in \cite{he2016identity,he2016deep}; 2) learning the transformation makes the explanation more user-friendly (more details in experiment section).

\begin{equation}
    \mathcal{L}_{adv} = \mathbbm{E}_{\mathbf{x}}[\log D(\mathbf{x})] + \mathbbm{E}_{\mathbf{x}, y^c}[\log(1 -  D(\mathbf{x}+G(\mathbf{x}, y^c)))]
\end{equation}

where $G$ generates an image perturbation $G(\mathbf{x}, y^c)$ conditioned on both the input image $\mathbf{x}$ and the target domain label $y^c$, and $D$ is the discriminator.

\noindent\textbf{Domain classification loss} 
Because we aim to render the generate image $\mathbf{x} + G(\mathbf{x}, y^c)$ conditional on $y^c$, an auxiliary classifier ($D_a$) is added on top of $D$ that imposes the domain classification loss when optimizing both $D$ and $G$. For the real and fake images, the loss is implemented by

\begin{equation}
    \mathcal{L}^r_{cls} = \mathbbm{E}_{\mathbf{x}, y}[-\log  D_a(y|\mathbf{x})]
\end{equation}
\begin{equation}
    \mathcal{L}^f_{cls} = \mathbbm{E}_{\mathbf{x}, y^c}[-\log  D_a(y^c|\mathbf{x}+G(\mathbf{x}, y^c))]
\label{equ:do_f}
\end{equation}

\noindent\textbf{Reconstruction loss} For counterfactual explanation, we do not hope the translated image is obtained by changing a large spatial region of the original. In an extreme case, erasing the original and drawing a new of counter class $y^c$. This will be no difference with only presenting an image of counter class as feedback (In this case, no explanation is provided). In other words, we hope that translated images preserve the content of its input images while changing only the domain-related part of the inputs. To alleviate this problem, we apply a cycle consistency loss~\cite{zhu2017unpaired} to the generator, defined as

\begin{equation}
    \mathcal{L}_{rec} = \mathbbm{E}_{\mathbf{x}, y^c, y}[||\mathbf{x} - (\mathbf{x}+G(\mathbf{x}, y^c) + G(\mathbf{x}+G(\mathbf{x}, y^c), y)) ||_1]
\end{equation}

\noindent\textbf{Explanation loss} Because our explanation is post-hoc, we add a pre-trained classifier $H$ to the end of the $G$, where $H$ is fixed through the entire training process. This is to guarantee the produced fake image $\mathbf{x}+G(\mathbf{x}, y^c)$ can fit the distribution of $H$. 

\begin{equation}
    \mathcal{L}_{exp} = \mathbbm{E}_{\mathbf{x}, y^c}[-\log  H(y^c|\mathbf{x}+G(\mathbf{x}, y^c))]
\end{equation}

It should note that although the loss seems to have the same effect with (\ref{equ:do_f}), their objectives are different. The goal of the latter is to generally optimize the output to the target domain, while that of the former is to guarantee the output is to explain the given classifier. One concern is that whether the explanation loss will lead to the generation of adversarial examples. We empirically found it did not happen probably because of two reasons: 1) the generator forces the produced image subject to the natural image distribution which is different from that of adversarial examples; 2) the discriminator make the fooling to the classifier $H$ difficult~\cite{song2018constructing,xiao2018generating}.

\noindent\textbf{Perturbation loss} In order to further constrain the perturbation is small, apart from reconstruction loss, a perturbation loss is added to directly force this. The former can be seen as an indirect way. It should be noted that the design of perturbation loss is based on a natural assumption that images of the counterfactual class should close to those of the prediction class in perceptual distance~\cite{johnson2016perceptual,zhang2018unreasonable}. For example, people are unlikely thinking a counter class ``house'' for a ``dog'' image. Consequently, the perturbation is small and a regularization, on the output of the generator, is requisite. Without this, the generator will tend to perturb the whole query image. This goes away from explanations.

\begin{equation}
    \mathcal{L}_{per} = \mathbbm{E}_{\mathbf{x}, y^c, y}[||G(\mathbf{x}, y^c) ||_1 + || G(\mathbf{x}+G(\mathbf{x}, y^c), y))||_1]
\end{equation}

\noindent\textbf{Final loss} The final objective functions to optimize $G$ and $D$ are summarized below,

\begin{equation}
    \mathcal{L}_D =  - \mathcal{L}_{adv} + \lambda_{cls} \mathcal{L}^r_{cls}
\end{equation}

\begin{equation}
    \mathcal{L}_G =  \mathcal{L}_{adv} + \lambda_{cls} \mathcal{L}^f_{cls} + \lambda_{rec} \mathcal{L}_{rec} + \lambda_{exp} \mathcal{L}_{exp} + \lambda_{per} \mathcal{L}_{per}
\end{equation}

where $\mathcal{L}_{adv}, \lambda_{cls}, \lambda_{rec}, \lambda_{exp}, \lambda_{per}$ are hyper-parameters that control the relative importance of each component. FRACE is optimization free in the sense that after the training stage, the query image can be fed into the model to get a counterfactual explanation by forwarding. This is different from the current exhaustive search in feature space~\cite{goyal2019counterfactual} and attacks~\cite{dhurandhar2018explanations} where the optimization has to be conducted for each query image. This makes our proposal much fast and applicable to real-time applications.

\begin{figure}[t]
\vskip 0.2in
\begin{center}
\centerline{\includegraphics[width=\columnwidth]{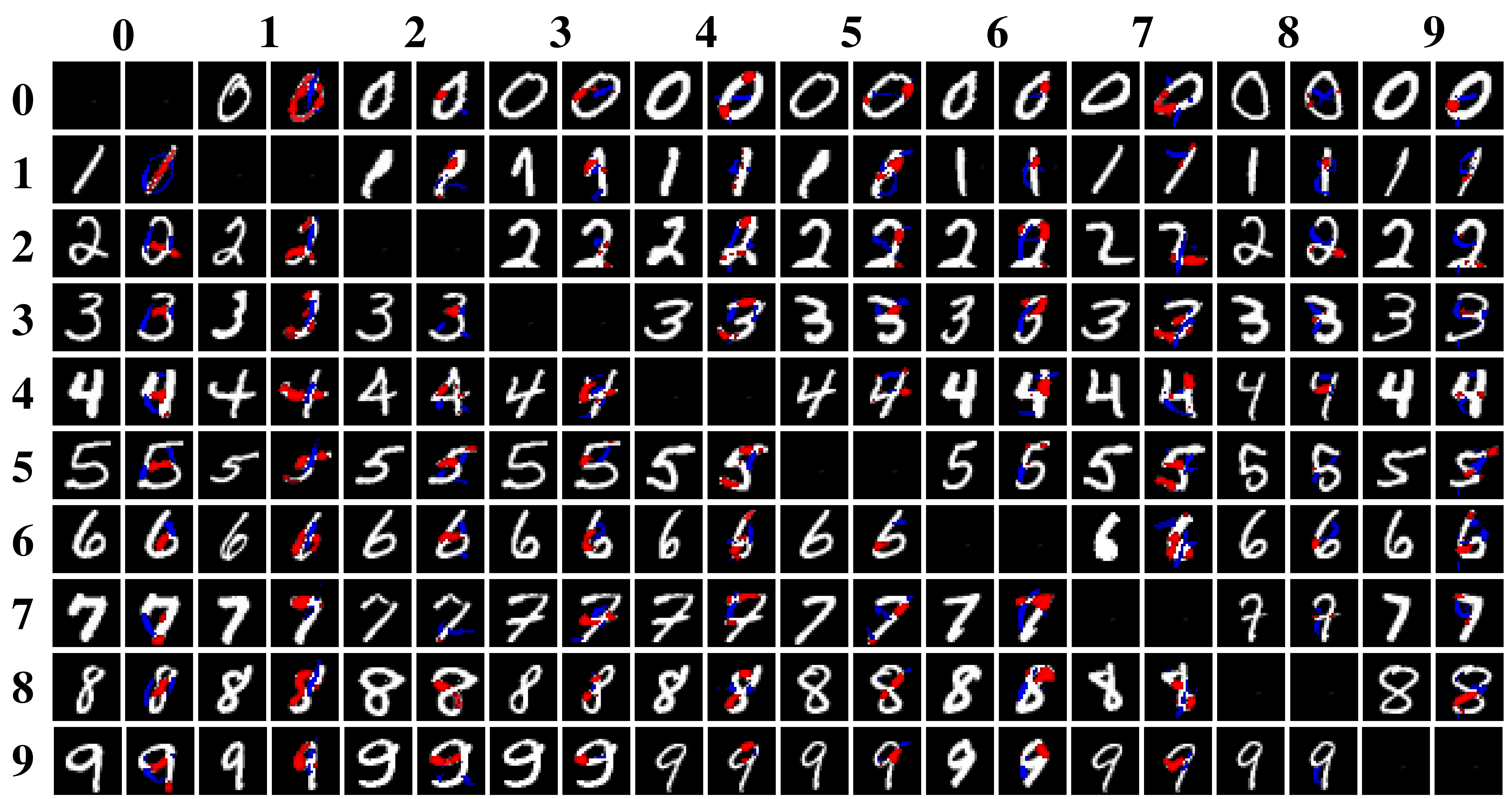}}
\caption{Explanations on MNIST for $100$ randomly selected combinations of prediction and counter classes.}
\label{fig:mnist}
\end{center}
\vskip -0.2in
\end{figure}

\begin{figure}[t]
\vskip 0.2in
\begin{center}
\centerline{\includegraphics[width=\columnwidth]{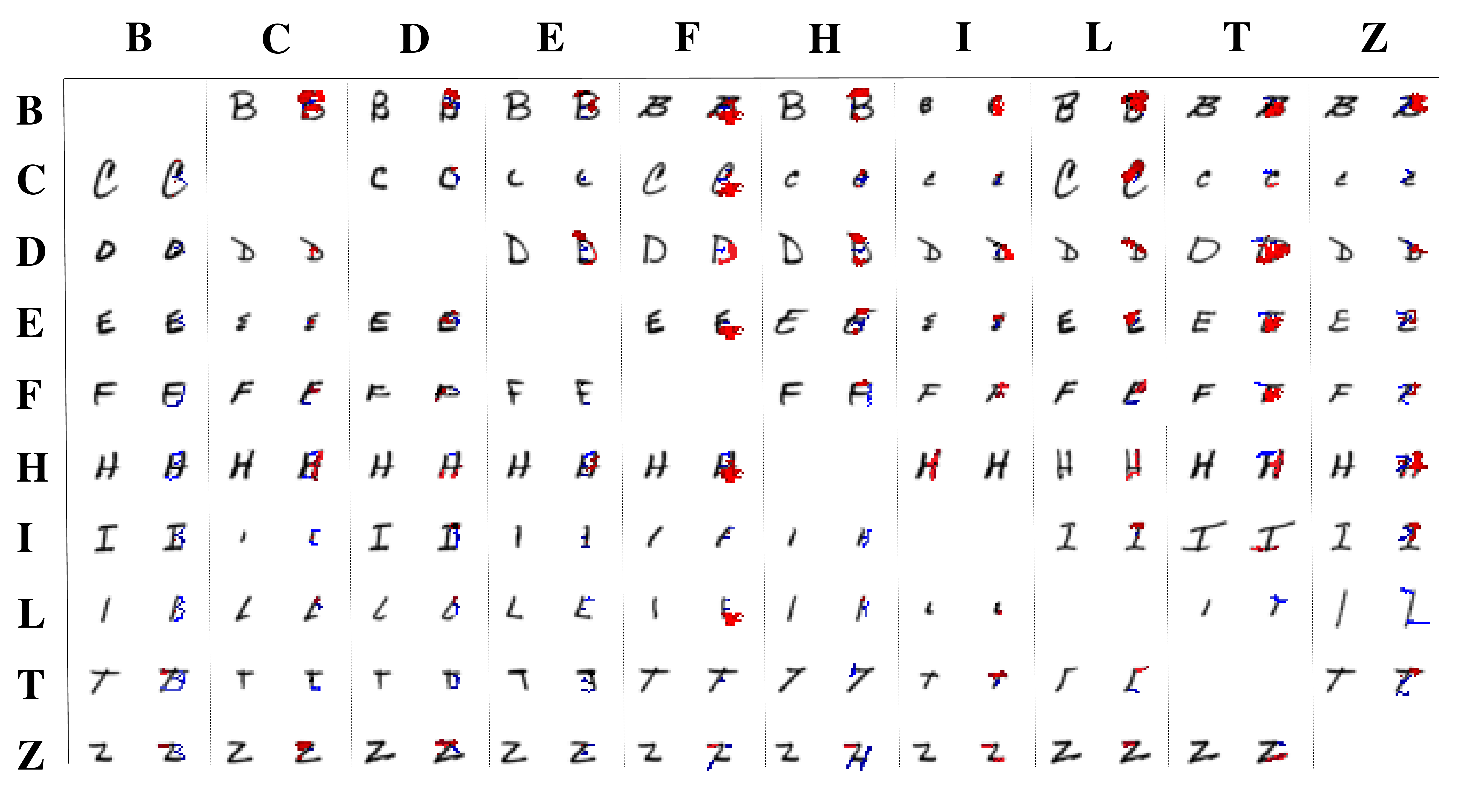}}
\caption{Explanations on EMNIST for $100$ randomly selected combinations of prediction and counter classes.}
\label{fig:letter}
\end{center}
\vskip -0.2in
\end{figure}

\begin{figure}[t]
\vskip 0.2in
\begin{center}
\centerline{\includegraphics[width=0.95\columnwidth]{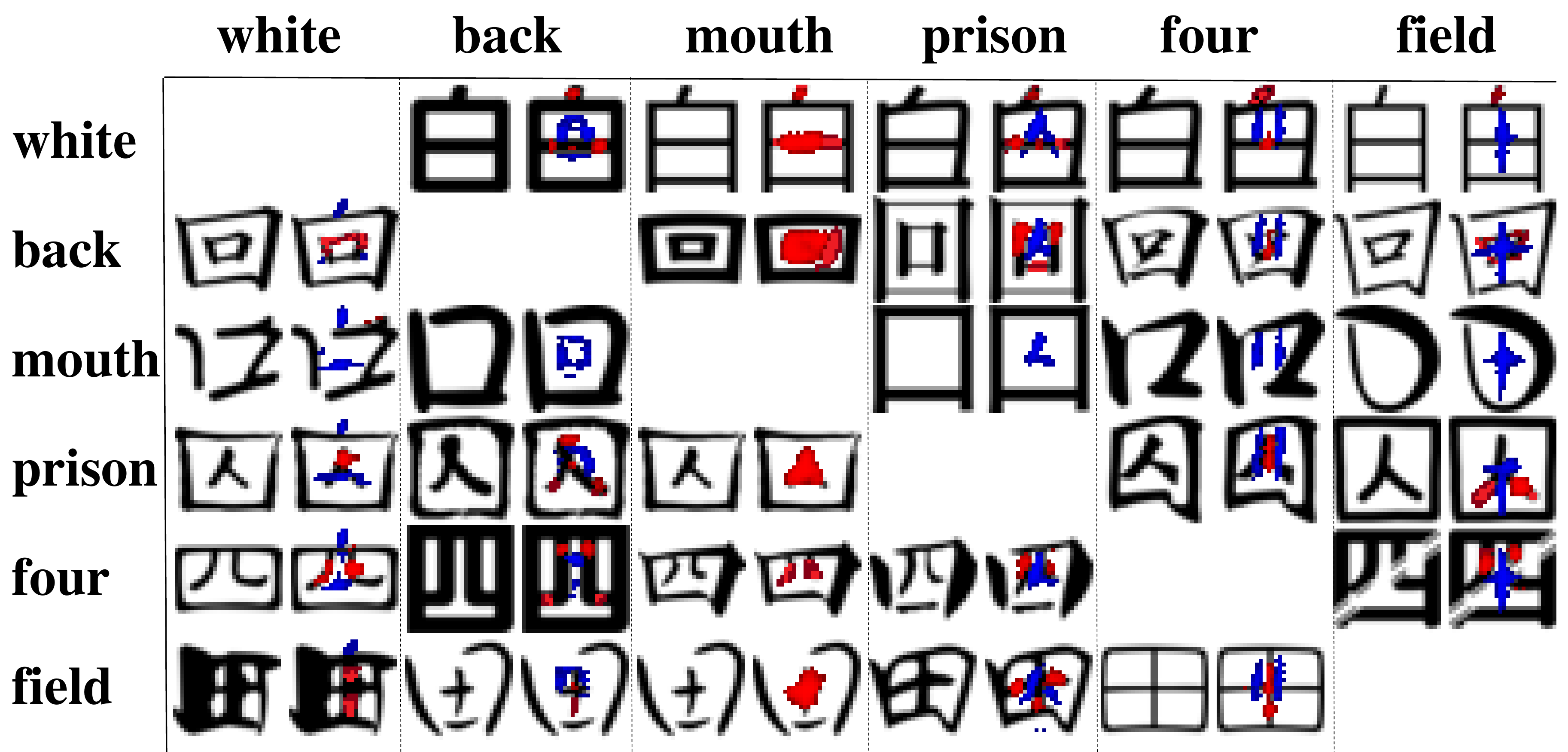}}
\caption{Explanations on Chinese Character for $36$ randomly selected combinations of prediction and counter classes.}
\label{fig:chinese}
\end{center}
\vskip -0.2in
\end{figure}

\begin{table*}[t]
\scriptsize
    \centering
\begin{tabular}{lcccc}
\toprule
Algorithms & Search~\cite{goyal2019counterfactual} & SCOUT~\cite{wang2020scout} & CEM~\cite{dhurandhar2018explanations} & Ours\\
\midrule
IPS & 6.15 $\pm$ 0.73& 84.62 $\pm$ 17.07& 63.43 $\pm$ 13.85& {\bf 201.44} $\pm$ 28.56\\
\bottomrule
\end{tabular}
    \caption{Speed comparison across different methods. IPS: images per second, implemented on NVIDIA TITAN Xp. Results are shown as mean $\pm$ stddev}
    \label{tab:comparison}
\end{table*}

\section{Experiments}

\noindent\textbf{Datasets} Three datasets with various difficulties were used to evaluate our proposal. {\bf MNIST}~\cite{lecun1998gradient} is a basic handwritten digits dataset. The capital letters extracted from EMNIST Letters~\cite{emnist} were assembled as our second dataset, named ``{\bf Letter}'' for simplicity. In addition, we collected a new Chinese Character ({\bf CC}) dataset from \cite{chinese_char}. The dataset consists of $10$ closely similar Chinese character categories. For each category, training size ranges from $2000$ to $3000$ examples. This dataset is fine-grained because of its large intra-class diversity and small inter-class differences. Some examples are shown in Figure \ref{fig:chinese_example}.

\noindent\textbf{Implementation details} For Letter and CC, $10\%$ of the examples were randomly selected as the testing set. For all datasets, ResNet-18~\cite{he2016deep} was used for the classifier. It is trained by $80$ epochs with SGD. Learning rate is $0.1$ at the beginning and degrades to $0.1$ times at the $40^{th}$, $60^{th}$ epoch. Weight decay is $0.0001$. The architecture and training details of FRACE fully follows \cite{choi2018stargan} and $\mathcal{L}_{adv}, \lambda_{cls}, \lambda_{rec}, \lambda_{exp}, \lambda_{per}$ were all set as $1$ in all our experiments. 

\subsection{Qualitative results}

We first present our explanations on three datasets in Figure \ref{fig:mnist}, \ref{fig:letter}, \ref{fig:chinese}. The row index denotes the prediction classes and column index counterfactual classes. For each prediction-counter class pair, two images are shown: query image and our explanation. The explanation is generated by the query image overlapped by the output of the generator $G(\mathbf{x}, y^c)$. The blue markings represent the positive values of $G(\mathbf{x}, y^c)$ and red markings represent the negative values, which means in order to transfer the query images to an image of counterfactual class, the red regions of the former should be erased, whereas the blue regions should be added. This constructs our explanation as ``if erasing the red regions and adding blue, the image would belong to the counterfactual class.'' On the results, it can be seen that our explanation is realistic and sensible across all prediction-counterfactual class pairs.

\subsection{Comparison to state of the art}

\begin{figure}[t]
\vskip 0.2in
\begin{center}
\centerline{\includegraphics[width=0.8\columnwidth]{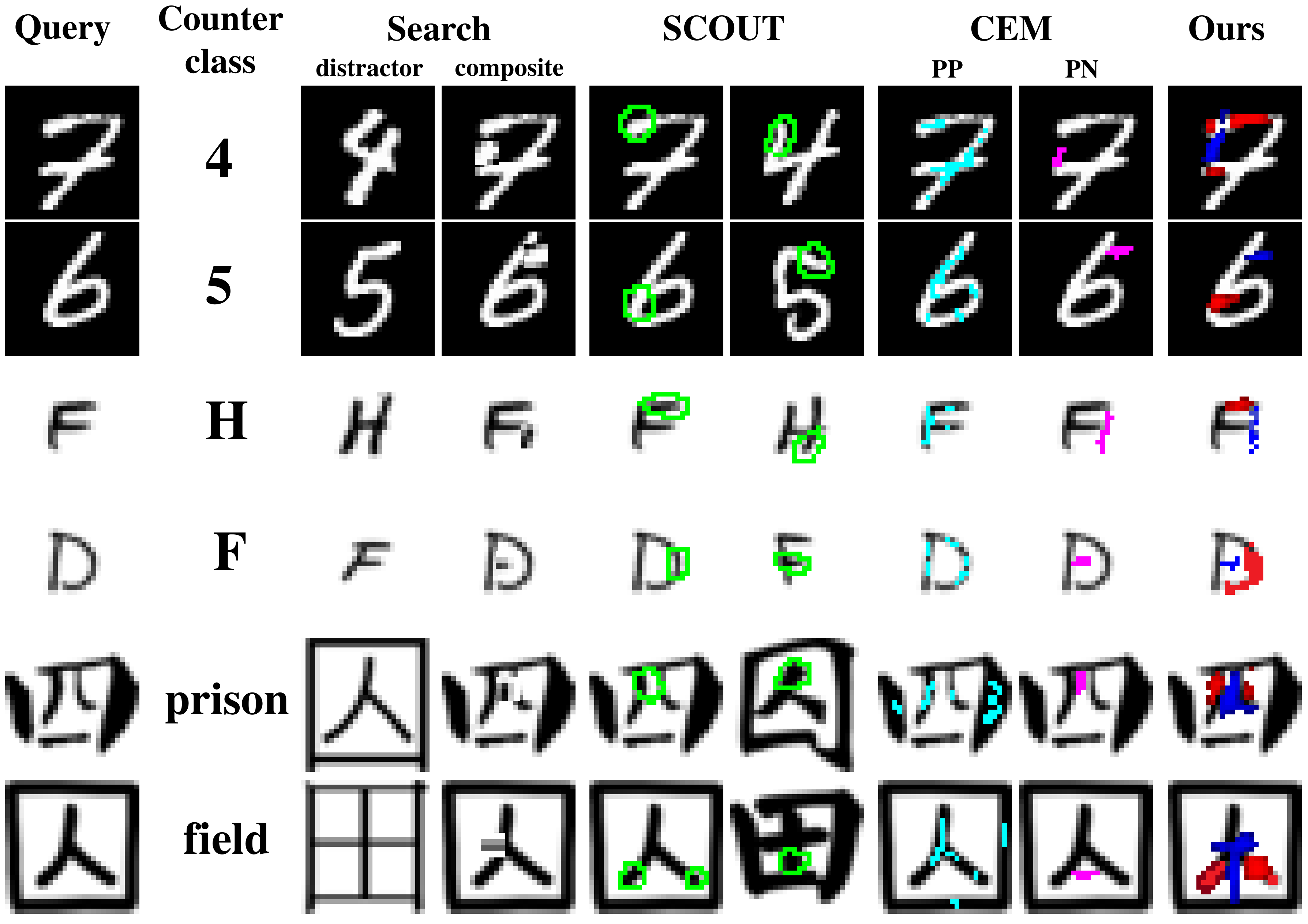}}
\caption{Explanation comparisons of our proposal with three state-of-the-art methods.}
\label{fig:comparison}
\end{center}
\vskip -0.2in
\end{figure}

Explanation methods are always hard to evaluate and compare because there is no ground truth is available. Although some previous works only show visualizations~\cite{shrikumar2017learning,sundararajan2017axiomatic,wang2019deliberative}, we compare our FRACE to state of the art qualitatively and quantitatively. Three state-of-the-art algorithms are considered. \cite{goyal2019counterfactual} first randomly selected a distractor image from the counter class, and then exhaustively searched a region on it so as to change the prediction class of the query image by replacing a certain region on the latter. We call this method ``Search'' for brevity in our paper. SCOUT~\cite{wang2020scout} also randomly selected an image from the counter class, but it only highlighted the discriminant regions on both the query and selected images by attribution based algorithms. CEM~\cite{dhurandhar2018explanations} is mostly similar to ours, but it produces two images for explanations. PP highlighted regions that have positive evidence for the prediction, while PN added regions that have negative effects on the prediction regarding the counter class. The qualitative comparison on several randomly selected examples is shown in Figure \ref{fig:comparison}. For the digit ``7'' with counter class ``4'' example, both Search and CEM could find a vertical line that should be added at the left above the middle horizontal, but only our proposal could also detect that the top horizontal line should be removed. Although SCOUT could also localize the class-specific features for both ``7'' and ``4'', its explanation was not straightforward and easily understandable compared to ours. From the comparison, it can be observed that our explanation is more realistic. Another merit of FRACE is that different from Search and SCOUT which both largely rely on the quality of the randomly selected images, our FRACE does not depend on such image.

We also compare the speed of producing the explanation across different algorithms. In Table \ref{tab:comparison}, it shows that our method is the fastest. This benefits from our forward mechanism. We only need one time forward to generate an explanation. Compared with ours, Search needs exhaustive search, making it the slowest one. CEM also needs optimization but it can use gradient descent to accelerate. Although SCOUT is also optimization-free, it relies on one time forward and back-propagation, as well as many post-computations. It should be noted that the speed comparison between ours and optimization-based ones is not unfair although GAN training is needed for FRACE, because the inference time cost is usually the main concern in real application. In addition, the same comparison way, optimization-based methods vs forward based methods, has been widely adopt in other literature~\cite{johnson2016perceptual,huang2017arbitrary}.  

\section{Discussion and conclusion}

In this work, we have proposed a novel fast real-time counterfactual explanation algorithm benefiting from the advanced properties of GAN. The explanation is based on a realistic transformation from the query to counterfactual class. Because it is optimization-free, its speed is much fast. The adversarial training renders the generated images more natural. In experiments, we have demonstrated its effectiveness and efficiency on three datasets, especially in a hard fine-grained expert domain set. Admittedly, we tested our method on more complicated situations such as fine-grained bird recognition and found that it underperformed in comparison to what our presented.
This is not surprising because multi-domain to multi-domain image to image translation is always a hard problem in real-world situations. There is no existing method that works well on this task, but our proposal can benefit from future works.

\footnotesize
\bibliographystyle{plain}
\bibliography{referencebib}

\begin{thebibliography}{10}

\bibitem{chinese_char}
Peter Burkimsher.
\newblock
  https://blog.usejournal.com/making-of-a-chinese-characters-dataset-92d4065cc7cc.

\bibitem{choi2018stargan}
Yunjey Choi, Minje Choi, Munyoung Kim, Jung-Woo Ha, Sunghun Kim, and Jaegul
  Choo.
\newblock Stargan: Unified generative adversarial networks for multi-domain
  image-to-image translation.
\newblock In {\em Proceedings of the IEEE conference on computer vision and
  pattern recognition}, pages 8789--8797, 2018.

\bibitem{emnist}
Gregory Cohen, Saeed Afshar, Jonathan Tapson, and Andr{\'{e}} van Schaik.
\newblock {EMNIST:} an extension of {MNIST} to handwritten letters.
\newblock {\em CoRR}, abs/1702.05373, 2017.

\bibitem{dhurandhar2018explanations}
Amit Dhurandhar, Pin-Yu Chen, Ronny Luss, Chun-Chen Tu, Paishun Ting,
  Karthikeyan Shanmugam, and Payel Das.
\newblock Explanations based on the missing: Towards contrastive explanations
  with pertinent negatives.
\newblock In {\em Advances in neural information processing systems}, pages
  592--603, 2018.

\bibitem{goodfellow2014generative}
Ian Goodfellow, Jean Pouget-Abadie, Mehdi Mirza, Bing Xu, David Warde-Farley,
  Sherjil Ozair, Aaron Courville, and Yoshua Bengio.
\newblock Generative adversarial nets.
\newblock In {\em Advances in neural information processing systems}, pages
  2672--2680, 2014.

\bibitem{goyal2019counterfactual}
Yash Goyal, Ziyan Wu, Jan Ernst, Dhruv Batra, Devi Parikh, and Stefan Lee.
\newblock Counterfactual visual explanations.
\newblock {\em arXiv preprint arXiv:1904.07451}, 2019.

\bibitem{he2016deep}
Kaiming He, Xiangyu Zhang, Shaoqing Ren, and Jian Sun.
\newblock Deep residual learning for image recognition.
\newblock In {\em Proceedings of the IEEE conference on computer vision and
  pattern recognition}, pages 770--778, 2016.

\bibitem{he2016identity}
Kaiming He, Xiangyu Zhang, Shaoqing Ren, and Jian Sun.
\newblock Identity mappings in deep residual networks.
\newblock In {\em European conference on computer vision}, pages 630--645.
  Springer, 2016.

\bibitem{huang2017arbitrary}
Xun Huang and Serge Belongie.
\newblock Arbitrary style transfer in real-time with adaptive instance
  normalization.
\newblock In {\em Proceedings of the IEEE International Conference on Computer
  Vision}, pages 1501--1510, 2017.

\bibitem{johnson2016perceptual}
Justin Johnson, Alexandre Alahi, and Li~Fei-Fei.
\newblock Perceptual losses for real-time style transfer and super-resolution.
\newblock In {\em European conference on computer vision}, pages 694--711.
  Springer, 2016.

\bibitem{lecun1998gradient}
Yann LeCun, L{\'e}on Bottou, Yoshua Bengio, and Patrick Haffner.
\newblock Gradient-based learning applied to document recognition.
\newblock {\em Proceedings of the IEEE}, 86(11):2278--2324, 1998.

\bibitem{mirza2014conditional}
Mehdi Mirza and Simon Osindero.
\newblock Conditional generative adversarial nets.
\newblock {\em arXiv preprint arXiv:1411.1784}, 2014.

\bibitem{odena2017conditional}
Augustus Odena, Christopher Olah, and Jonathon Shlens.
\newblock Conditional image synthesis with auxiliary classifier gans.
\newblock In {\em Proceedings of the 34th International Conference on Machine
  Learning-Volume 70}, pages 2642--2651. JMLR. org, 2017.

\bibitem{ren2015faster}
Shaoqing Ren, Kaiming He, Ross Girshick, and Jian Sun.
\newblock Faster r-cnn: Towards real-time object detection with region proposal
  networks.
\newblock In {\em Advances in neural information processing systems}, pages
  91--99, 2015.

\bibitem{selvaraju2017grad}
Ramprasaath~R Selvaraju, Michael Cogswell, Abhishek Das, Ramakrishna Vedantam,
  Devi Parikh, and Dhruv Batra.
\newblock Grad-cam: Visual explanations from deep networks via gradient-based
  localization.
\newblock In {\em Proceedings of the IEEE international conference on computer
  vision}, pages 618--626, 2017.

\bibitem{shrikumar2017learning}
Avanti Shrikumar, Peyton Greenside, and Anshul Kundaje.
\newblock Learning important features through propagating activation
  differences.
\newblock In {\em Proceedings of the 34th International Conference on Machine
  Learning-Volume 70}, pages 3145--3153. JMLR. org, 2017.

\bibitem{song2018constructing}
Yang Song, Rui Shu, Nate Kushman, and Stefano Ermon.
\newblock Constructing unrestricted adversarial examples with generative
  models.
\newblock In {\em Advances in Neural Information Processing Systems}, pages
  8312--8323, 2018.

\bibitem{sundararajan2017axiomatic}
Mukund Sundararajan, Ankur Taly, and Qiqi Yan.
\newblock Axiomatic attribution for deep networks.
\newblock In {\em Proceedings of the 34th International Conference on Machine
  Learning-Volume 70}, pages 3319--3328. JMLR. org, 2017.

\bibitem{wang2019deliberative}
Pei Wang and Nuno Nvasconcelos.
\newblock Deliberative explanations: visualizing network insecurities.
\newblock In {\em Advances in Neural Information Processing Systems}, pages
  1374--1385, 2019.

\bibitem{Wang_2018_ECCV}
Pei Wang and Nuno Vasconcelos.
\newblock Towards realistic predictors.
\newblock In {\em The European Conference on Computer Vision (ECCV)}, September
  2018.

\bibitem{wang2020scout}
Pei Wang and Nuno Vasconcelos.
\newblock Scout: Self-aware discriminant counterfactual explanations.
\newblock {\em IEEE Conference on Computer Vision and Pattern Recognition},
  2020.

\bibitem{wang2017idk}
Xin Wang, Yujia Luo, Daniel Crankshaw, Alexey Tumanov, Fisher Yu, and Joseph~E
  Gonzalez.
\newblock Idk cascades: Fast deep learning by learning not to overthink.
\newblock {\em arXiv preprint arXiv:1706.00885}, 2017.

\bibitem{xiao2018generating}
Chaowei Xiao, Bo~Li, Jun-Yan Zhu, Warren He, Mingyan Liu, and Dawn Song.
\newblock Generating adversarial examples with adversarial networks.
\newblock {\em arXiv preprint arXiv:1801.02610}, 2018.

\bibitem{zhang2018unreasonable}
Richard Zhang, Phillip Isola, Alexei~A Efros, Eli Shechtman, and Oliver Wang.
\newblock The unreasonable effectiveness of deep features as a perceptual
  metric.
\newblock In {\em Proceedings of the IEEE conference on computer vision and
  pattern recognition}, pages 586--595, 2018.

\bibitem{zhu2017unpaired}
Jun-Yan Zhu, Taesung Park, Phillip Isola, and Alexei~A Efros.
\newblock Unpaired image-to-image translation using cycle-consistent
  adversarial networks.
\newblock In {\em Proceedings of the IEEE international conference on computer
  vision}, pages 2223--2232, 2017.

\bibitem{zhu2018overview}
Xiaojin Zhu, Adish Singla, Sandra Zilles, and Anna~N Rafferty.
\newblock An overview of machine teaching.
\newblock {\em arXiv preprint arXiv:1801.05927}, 2018.

\end{thebibliography}

\end{document}